\documentclass{article}

\PassOptionsToPackage{numbers, compress}{natbib}

\usepackage{floatrow}

\usepackage[preprint]{neurips_2022}

\usepackage[utf8]{inputenc} 
\usepackage[T1]{fontenc}    
\usepackage{hyperref}       
\usepackage{url}            
\usepackage{booktabs}       
\usepackage{amsfonts}       
\usepackage{amsmath}
\usepackage{amssymb}
\usepackage{mathtools}
\usepackage{nicefrac}       
\usepackage{microtype}      
\usepackage{xcolor}         

\usepackage{bm}
\usepackage{graphicx}
\usepackage{caption}
\usepackage{subcaption}
\usepackage{wrapfig}
\usepackage{authblk}
\graphicspath{ {./figs/} }

\usepackage{mathrsfs}
\usepackage{dsfont}

\begin{document}

\title{PropertyDAG: Multi-objective Bayesian optimization of partially ordered, mixed-variable properties for biological sequence design}

%


\author[1]{\textbf{Ji Won Park}}
\author[1]{\textbf{Samuel Stanton}}
\author[1]{\textbf{Saeed Saremi}}
\author[1]{\textbf{Andrew Watkins}}
\author[1]{\textbf{Henri Dwyer}}
\author[1]{\textbf{Vladimir Gligorijevi\'c}}
\author[1]{\textbf{Richard Bonneau}}
\author[1]{\textbf{Stephen Ra}}
\author[1,2,3,4]{\textbf{Kyunghyun Cho}}

\affil[1]{\footnotesize Prescient Design, Genentech}
\affil[2]{\footnotesize Department of Computer Science, Courant Institute of Mathematical Sciences, New York University}
\affil[3]{\footnotesize Center for Data Science, New York University}
\affil[4]{\footnotesize CIFAR Fellow}
\affil[ ]{\texttt{park.ji\_won@gene.com}}


\maketitle

\begin{abstract}
Bayesian optimization offers a sample-efficient framework for navigating the exploration-exploitation trade-off in the vast design space of biological sequences.
Whereas it is possible to optimize the various properties of interest jointly using a multi-objective acquisition function, such as the expected hypervolume improvement (EHVI), this approach does not account for objectives with a hierarchical dependency structure. 
We consider a common use case where some regions of the Pareto frontier are prioritized over others according to a specified $\textit{partial ordering}$ in the objectives. 
For instance, when designing antibodies, we maximize the binding affinity to a target antigen only if it can be expressed in live cell culture---modeling the experimental dependency in which affinity can only be measured for antibodies that can be expressed and thus produced in viable quantities. 
In general, we may want to confer a partial ordering to the properties such that each is optimized conditioned on its parent properties satisfying some feasibility condition. 
To this end, we present PropertyDAG, a framework that operates on top of the traditional multi-objective BO to impose this desired ordering on the objectives, e.g. expression $\rightarrow$ affinity. 
We demonstrate its performance over multiple simulated active learning iterations on a penicillin production task, toy numerical problem, and a real-world antibody design task.
\end{abstract}

\section{Introduction} \label{sec:intro}

Designing biological sequences entails searching over vast combinatorial design spaces. Recently, deep sequence generation models trained on large datasets of known, functional sequences have shown promise in generating physically and chemically plausible designs \citep[e.g.][]{biswas2021low,das2021accelerated,gligorijevic2021function}. Whereas these models accelerate the design process, limited resources place a cap on how many designs we can characterize \textit{in vitro} for assessing their suitability. Only once a design is validated \textit{in vitro} and undergoes multiple rounds of optimization can it proceed down the drug development pipeline to preclinical development and clinical trials, where its performance is tested \textit{in vivo}. 

Because the wet lab cannot provide feedback on all of the candidate designs, we take an iterative, data-driven approach to select the the most informative subset to submit to the wet lab. Many drug design applications call for such an active learning approach, as the initial datasets available to train predictive models on our desired properties of interest tends to be small or nonexistent. The measurements returned by the lab in each iteration is appended to our training set and we update our models using the augmented dataset for the next iteration. 

The wet lab's measurement process can be viewed as a black-box function that is expensive to evaluate. In the context of identifying designs maximizing this function, Bayesian optimization (BO) emerges as a promising, sample-efficient framework that trades off exploration (evaluating highly uncertain designs) and exploitation (evaluating designs believed to carry the best properties) in a principled manner \citep{jones1998efficient}. It relies on a probabilistic surrogate model that infers the posterior distribution over the objectives and an acquisition function that assigns an expected utility value to each candidate. BO has been successfully applied to a variety of protein engineering applications \citep{pyzer2018bayesian,bellamy2022batched,stanton2022accelerating}. 

In particular, we cast our problem as multi-objective BO, where multiple objectives are jointly optimized. 
Our objectives originate from the molecular properties evaluated during \textit{in vitro} validation. 
This validation process involves producing the design, confirming its pharmacology, and evaluating whether it is active against a given drug target of interest. 
If found to be potent, the design is then assayed for \textit{developability} attributes---physicochemical properties that characterize the safety, delivery, and manufacturability \citep{jarasch2015developability}. 

The experimental process of \textit{in vitro} validation signifies a hierarchy among the objectives. 
Consider the property ``expression'' in the context of antibody design, for instance. 
A designed antibody candidate must first be expressed in live cell culture. 
If the level of expression does not meet a fixed threshold, the lab cannot produce it and it cannot be assayed for potency and developability downstream. Supposing now that a design did express in viable amounts, if it does not bind to a target antigen with sufficient ``affinity'' (and is thus not potent), then the design fails as an antibody and there is little practical incentive in assaying it for developability (such as specificity and thermostability)---even if it is possible to do so. 
The dependency between properties, whether experimental or biological in origin, motivates us prioritize some objectives before others when selecting the subset of designs to submit to the wet lab. 
Our primary goal is to identify ``joint positive'' designs, designs that meet the chosen thresholds in all the parent objectives (expressing binders) according to the specified partial ordering and also perform well in the leaf-level objectives (high specificity, thermostability). 

To this end, we propose PropertyDAG, a simple framework that operates on top of the traditional multi-objective BO to impose a desired partial ordering on the objectives, e.g. expression $\rightarrow$ affinity $\rightarrow$ $\{$ specificity, thermostability$\}$. Our framework modifies the posterior inference procedure within standard BO in two ways. First, we treat the objectives as \textit{mixed-variable}---in particular, each objective is modeled as a mixture of zeros and a wide dispersion of real-valued, non-zero values. The surrogate model consists of a binary classifier and a regressor, which infer the zero mode and the non-zero mode, respectively. We show that this modeling choice is well-suited for biological properties, which tend to carry excess zero, or null, values and fat tails \citep{jain2017biophysical}. Second, before samples from the posterior distribution inferred by the surrogate model enter the acquisition function, we transform the samples such that they conform to the specified partial ordering of properties. 
We run multi-objective BO with PropertyDAG over multiple simulated active learning iterations to a penicillin production task, a toy numerical problem, and a real-world antibody design task. In all three tasks, PropertyDAG-BO identifies significantly more joint positives compared to standard BO. After the final iteration, the surrogate models trained under Property-BO also output more accurate predictions on the joint positives in a held-out test set than do the standard BO equivalents. 

\section{Background} 
\label{sec:bg}

\textbf{Bayesian optimization} (BO) is a popular technique for sample-efficient black-box optimization \cite[see][for a review]{shahriari2015taking, frazier2018tutorial}. 
Suppose our objective $f: \mathcal{X} \rightarrow \mathbb{R}$ is a black-box function of the design space $\mathcal{X}$ that is expensive to evaluate. 
Our goal is to efficiently identify a design $\bm{x}^\star \in \mathcal{X}$ maximizing\footnote{
For simplicity, we define the task as maximization in this paper without loss of generality. 
For minimizing $f$, we can negate $f$, for instance.
} 
$f$. 
BO leverages two tools, a probabilistic surrogate model and a utility function, to trade off exploration (evaluating highly uncertain designs) and exploitation (evaluating designs believed to maximize $f$) in a principled manner. 

For each iteration $t \in \mathbb{N}$, we have a dataset $\mathcal{D}_t = \{( \bm{x}^{(1)}, {y}^{(1)} ), \cdots, ( \bm{x}^{(N_t)}, {y}^{(N_t)} )\} \in \mathscr{D}_t$, where each ${y}^{(n)}$ is a noisy observation of $f$.
First, the probabilistic model $\hat f: \mathcal{X} \rightarrow \mathbb{R}$ infers the posterior distribution $p(\hat f | \mathcal{D}_t)$, quantifying the plausibility of surrogate objectives $\hat f \in \mathcal{F}$. 
Next, we introduce a utility function $u: \mathcal{X} \times \mathcal{F} \times \mathscr{D}_t : \rightarrow \mathbb{R}$.
The acquisition function $a(\bm x)$ is simply the expected utility of $\bm x$ w.r.t. our beliefs about $f$,
\begin{align}
    a(\bm x) = \int u(\bm x, \hat f, \mathcal{D}_t)p(\hat f | \mathcal{D}_t) d\hat f.
\end{align}
For example, we obtain the expected improvement (EI) acquisition function if we take $ u_{\mathrm{EI}}(\bm x, \hat f, \mathcal{D}) = [\hat f(\bm x) - \max_{(\bm x', y') \in \mathcal{D}} y']_+,$ where $[\cdot]_+ = \max(\cdot, 0)$ \citep{movckus1975bayesian,jones1998efficient}. 
Generally the integral is approximated by Monte Carlo (MC) with posterior samples $\hat{f}^{(j)} \sim p ( \hat f|\mathcal{D}_t )$.
We select a maximizer of $a$ as the new design, measure its properties, and append the observation to the dataset. The surrogate is then retrained on the augmented dataset and the procedure repeats. 

\paragraph{Multi-objective optimization} 

When there are multiple objectives of interest, a single best design may not exist.
Suppose there are $K$ objectives, $f: \mathcal{X} \rightarrow \mathbb{R}^K$.
The goal of multi-objective optimization (MOO) is to identify the set of \textit{Pareto-optimal} solutions such that improving one objective within the set leads to worsening another. 
We say that $\bm x$ dominates $\bm x'$, or ${f}(\bm{x}) \succ {f}(\bm{x}')$, if $f_k(\bm{x}) \geq f_k(\bm{x}')$ for all $k \in \{1, \dotsc, K\}$ and $f_k(\bm x) > f_k(\bm x')$ for some $k$.
The set of \textit{non-dominated} solutions $\mathscr{X}^*$ is defined in terms of the Pareto frontier (PF) $\mathcal{P}^*$,
\begin{align} 
\label{eq:pareto}
\mathscr{X}^\star = \{\bm{x}: f(\bm{x}) \in \mathcal{P}^\star\}, \hspace{4mm} \text{where } \mathcal{P}^\star = \{f(\bm{x}) \: : \: \bm x \in \mathcal{X}, \;  \nexists \: \bm{x}' \in \mathcal{X} \textit{ s.t. } f(\bm{x}') \succ f(\bm{x}) \}.
\end{align}

 MOO algorithms typically aim to identify a finite approximation to $\mathscr{X}^\star$, which may be infinite, within a reasonable number of iterations.
One way to measure the quality of an approximate PF $\mathcal{P}$ is to compute the hypervolume ${\rm HV}(\mathcal{P} | \bm{r}_{\rm ref})$ of the polytope bounded by $\mathcal{P} \cup \{\bm r_{\mathrm{ref}}\}$, where $\bm r_{\mathrm{ref}} \in \mathbb{R}^K$ is a user-specified \textit{reference point}.
We obtain the expected hypervolume improvement (EHVI) acquisition function if we take
\begin{align} 
    u_{\mathrm{EHVI}}(\bm x, \hat f, \mathcal{D}) = {\rm HVI}(\mathcal{P}', \mathcal{P} | \bm{r}_{\rm ref}) = [{\rm HV}(\mathcal{P}' | \bm{r}_{\rm ref}) - {\rm HV}(\mathcal{P} | \bm{r}_{\rm ref})]_+, \label{eq:ehvi}
\end{align}
where $\mathcal{P}' = \mathcal{P} \cup \{\hat f(\bm x)\}$ \citep{emmerich2005single,emmerich2011hypervolume}.

 

\paragraph{Noisy observations}

In the noiseless setting, the observed baseline PF is the true baseline PF, i.e. $\mathcal{P}_t = \{\bm{y}: \bm{y} \in \mathcal{Y}_t, \: \nexists \: \bm{y}' \in \mathcal{Y}_t \textit{ s.t. } \bm{y}' \succ \bm{y} \}$ where $\mathcal{Y}_t \coloneqq \{\bm{y}^{(n)}\}_{n=1}^{N_t}$. This does not, however, hold in many practical applications, where measurements carry noise. For instance, given a zero-mean Gaussian measurement process with noise covariance $\Sigma$, the feedback for a design $\bm{x}$ is $\bm{y} \sim \mathcal{N}\left( {f}(\bm{x}), \Sigma \right)$, not $f(\bm{x})$ itself. 
The \textit{noisy} expected hypervolume improvement (NEHVI) acquisition function marginalizes over the surrogate posterior at the previously observed points $\mathcal{X}_t = \{\bm{x}^{(n)}\}_{n=1}^{N_t}$,
\begin{align} 
    u_{\mathrm{NEHVI}}(\bm x, \hat f, \mathcal{D}) = {\rm HVI}(\hat{\mathcal{P}}_t', \hat{\mathcal{P}}_t | \bm{r}_{\rm ref}) \label{eq:nehvi},
\end{align}
where $\hat{\mathcal{P}}_t = \{\hat f(\bm{x}) \: : \: \bm x \in \mathcal{X}_t, \;  \nexists \: \bm{x}' \in \mathcal{X}_t \textit{ s.t. } \hat f(\bm{x}') \succ \hat f(\bm{x}) \}$ and $\hat{\mathcal{P}}' = \hat{\mathcal{P}} \cup \{\hat{f}(\bm{x})\}$ \citep{daulton2021parallel}.

\paragraph{Batched (parallel) optimization} 

Sequential optimization, or querying $f$ for one design per iteration, is impractical for many applications due to the latency in feedback. In protein engineering, for example, it may be necessary to select a batch of designs in a given iteration and wait several months to receive measurements \citep{mayr2009novel,sinai2020primer}. Jointly selecting a batch of $q$ designs from a large pool of $q' \gg q$ candidates requires combinatorial evaluations of the acquisition function. 

\section{Related Work} \label{sec:related_work}

Existing work on multi-objective BO does not account for objectives with a hierarchical dependency structure \citep{gelbart2014bayesian,wada2019bayesian,yang2019multi,daulton2020differentiable,daulton2021parallel}. We refer to \cite{astudillo2021bayesian} for a formulation of single-objective BO with a hierarchy in how the objective is computed. A body of work focuses on constrained optimization, which optimizes a black-box function subject to a set of black-box constraints being satisfied \cite{wu2016parallel,gardner2014bayesian,hernandez2016general,ginsbourger2010kriging,letham2019constrained, malkomes2021beyond}. For dealing with mixed-variable objectives, \cite{daulton2022bayesian} propose reparameterizing the discrete random variables in terms of continuous parameters. Our approach here is to model them explicitly using the zero-inflated formalism.

\section{Method}
\label{sec:method}
\begin{figure}[ht] 
  \centering 
  \includegraphics[scale=0.75]{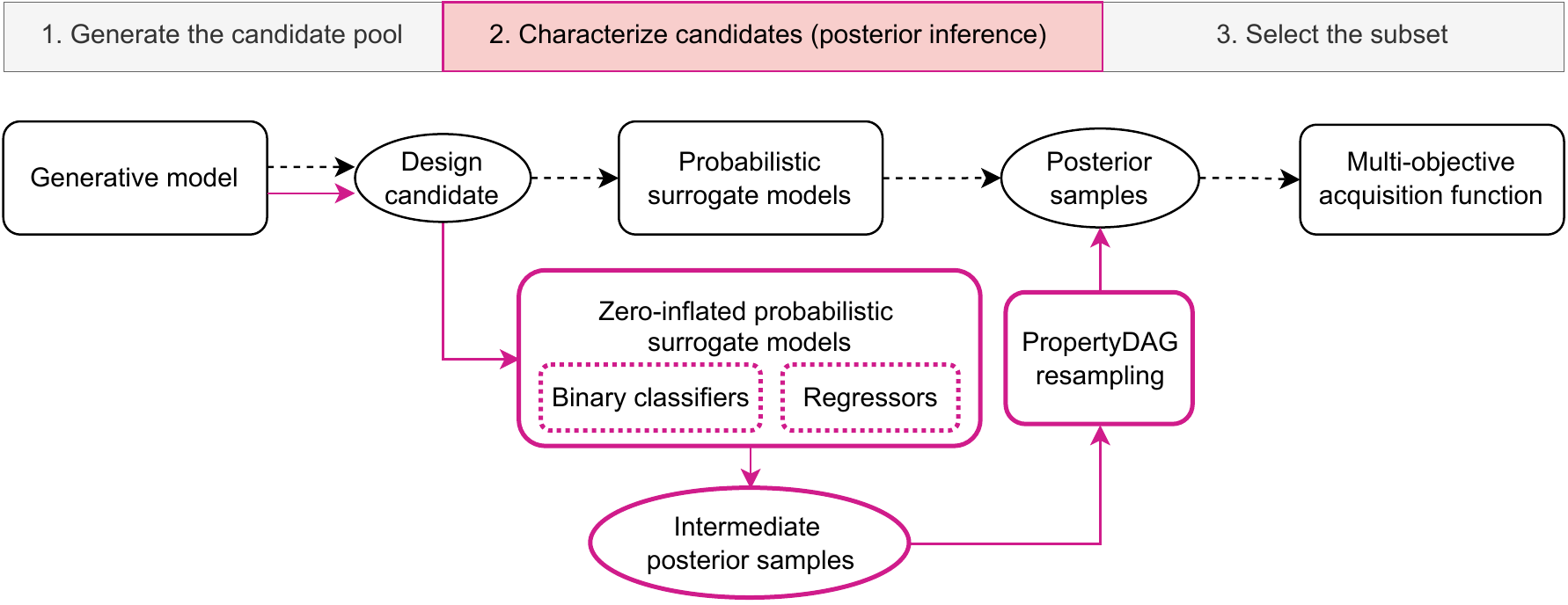} 
  \caption{Comparison of a traditional multi-objective BO pipeline (dashed black arrows) with the proposed pipeline enabled by PropertyDAG (solid magenta arrows). \label{fig:pipeline}}
\end{figure}

\paragraph{Overview} 
Figure~\ref{fig:pipeline} illustrates our proposed PropertyDAG-BO framework alongside standard BO. The candidate generation step is identical in both cases; we first sample a large pool of design candidates from a proposal distribution, often implemented as a generative model. The difference lies in the next step, in which the designs are scored by probabilistic surrogate models. In PropertyDAG-BO, we must first specify a partial ordering of our objectives (Section~\ref{sec:defining_property_dag}). We then explicitly assign a zero-inflated generative model for each objective (Section~\ref{sec:zero_inflation}) such that a probabilistic classifier models its ``zero'' mode and a probabilistic regressor models the remaining continuous-valued ``non-zero'' mode. The raw posterior samples from the surrogates then undergo a ``resampling'' step (Section~\ref{sec:resampling}) that enforces the specified PropertyDAG. Finally, the modified posterior samples enter the multi-objective acquisition function, which scores the design candidates just as in standard BO.

\subsection{Defining a PropertyDAG} \label{sec:defining_property_dag}

Many drug design applications motivate us to prescribe some hierarchy among our objective properties of interest. The partial ordering may arise from an experimental dependency, e.g. a design candidate must pass a certain threshold in one property for its other properties to be measured. In the context of antibody design, a design candidate is a sequence of amino acids representing an antibody that must first be expressed in cell culture. If the level of expression does not exceed some threshold in mass per volume, the lab cannot produce it in viable amounts and it cannot be assayed for other properties, such as binding affinity to a target antigen. Our PropertyDAG may then take the form: expression $\rightarrow$ affinity. Experimental dependencies like this creates an asymmetry among the objectives; it reduces the information content of designs that do not express much more than that of designs that do not bind, because non-expressing designs cannot provide binding measurements. 

Alternatively, the partial ordering may encode our preference for the types of designs we want to obtain. We may prioritize a property, for instance, so that we reject designs that performs poorly in this property, no matter how well they perform in all the others. If a designed antibody does not bind to the target antigen, it has failed in its primary function, so we have little interest in its developability properties, such as specificity to the target antigen and thermostability, even though, unlike for non-expressers, they often remain measurable. We then impose the following PropertyDAG: expression $\rightarrow$ affinity $\rightarrow$ $\{$ specificity, thermostability $\}$. 

\begin{figure}[t!]
\floatbox[{\capbeside\thisfloatsetup{capbesideposition={right, top},capbesidewidth=8cm}}]{figure}[\FBwidth]
{\caption{Example of a PropertyDAG, with three properties and two levels of hierarchy. Each property is denoted $y_{l, m}$, where $l$ indexes the level and $m$ indexes sibling properties at the same level. Magenta arrows refer to the dependencies imposed by PropertyDAG. Black arrows make it explicit that each $y_{l, m}$ is modeled as zero-inflated, where $b_{l, m}$ governs the zero events and $r_{l, m}$ governs the continuous non-zero events.}\label{fig:pgm}}
{\includegraphics[width=0.3\textwidth]{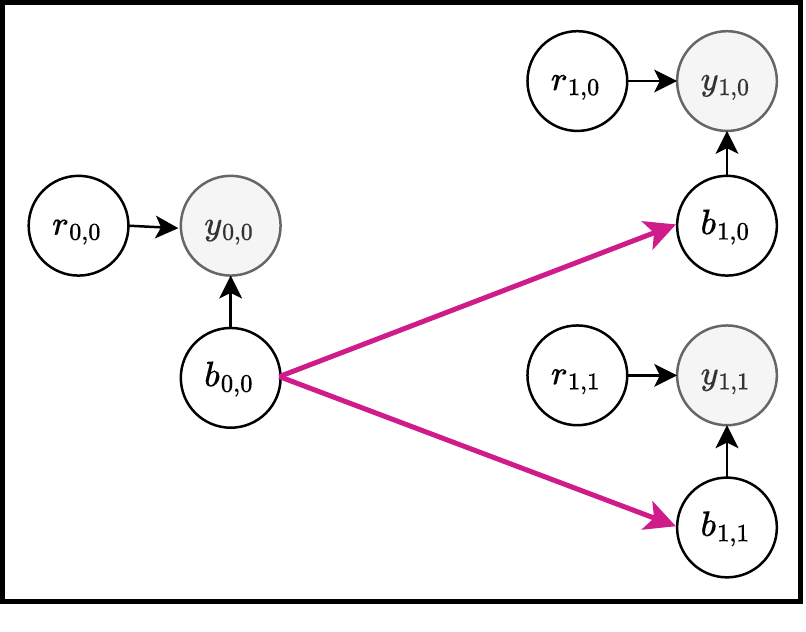}}
\end{figure}

A PropertyDAG can be expressed as ordered sets of properties: $\{y_{0, 0}, \dotsc, y_{0, M_0}\} \rightarrow \{y_{1, 0}, \dotsc, y_{1, M_1}\} \rightarrow \cdots \rightarrow \{y_{L, 0}, \dotsc, y_{L, M_L}\}$, where $y_{l, m}$ is the property at level $l \in \{0, \dotsc, L-1\}$ of the hierarchy and $m \in \{0, \dotsc, M_l-1\}$ is its index among the $M_l$ sibling properties at the same level $l$. Figure~\ref{fig:pgm} shows an example of a PropertyDAG with three properties and two levels.

\begin{figure}[b!] 
    \centering
    \begin{subfigure}[t]{0.45\textwidth}
        \centering
        \includegraphics[scale=0.48, trim=0.cm 6cm 0.cm 4cm]{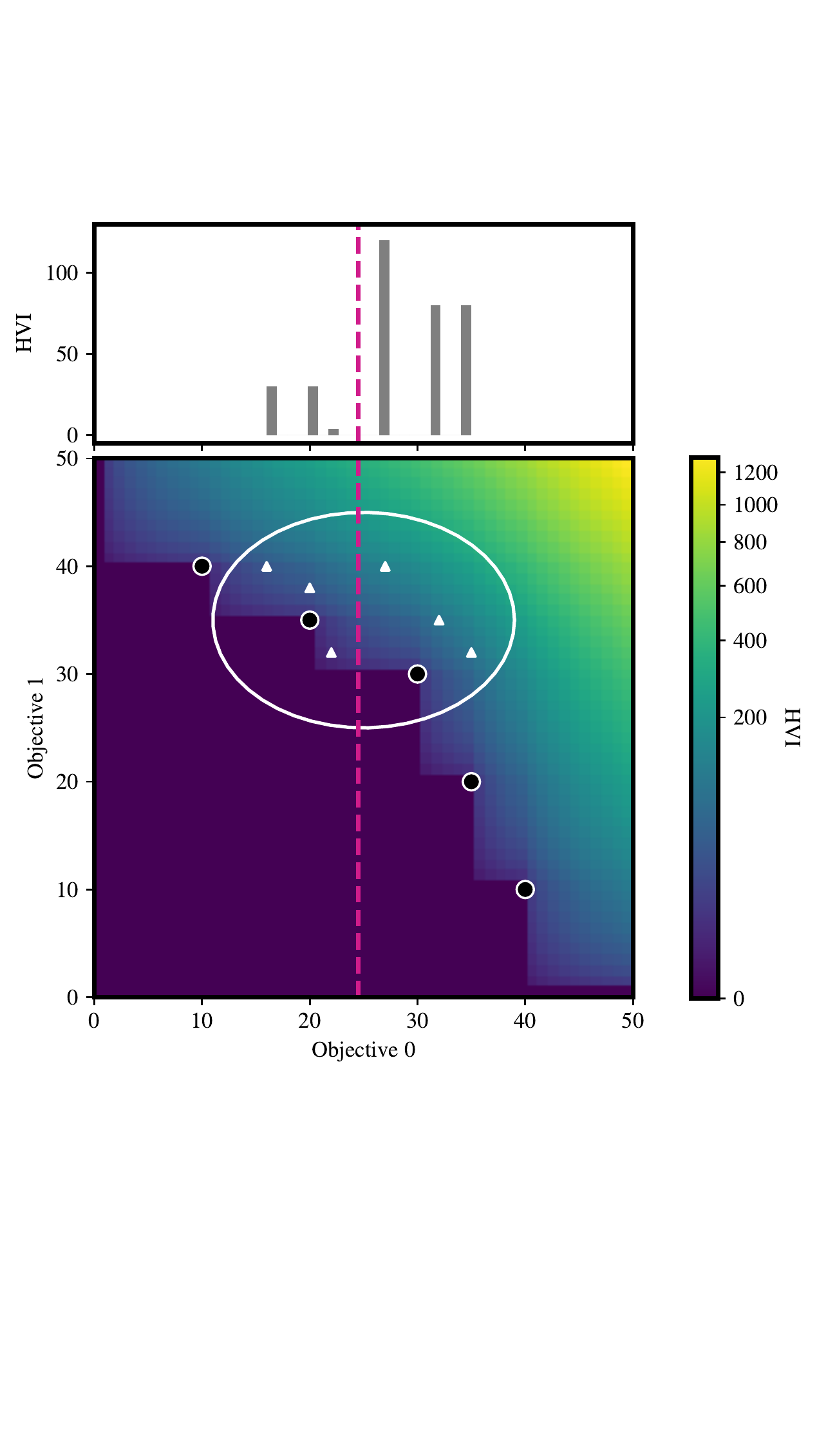}
        \caption{Default}
        \label{fig:pareto_schematic_default}
    \end{subfigure}
    \hfill
    \begin{subfigure}[t]{0.45\textwidth}
        \centering
        \includegraphics[width=\linewidth, trim=0.cm 6cm 0.cm 4cm]{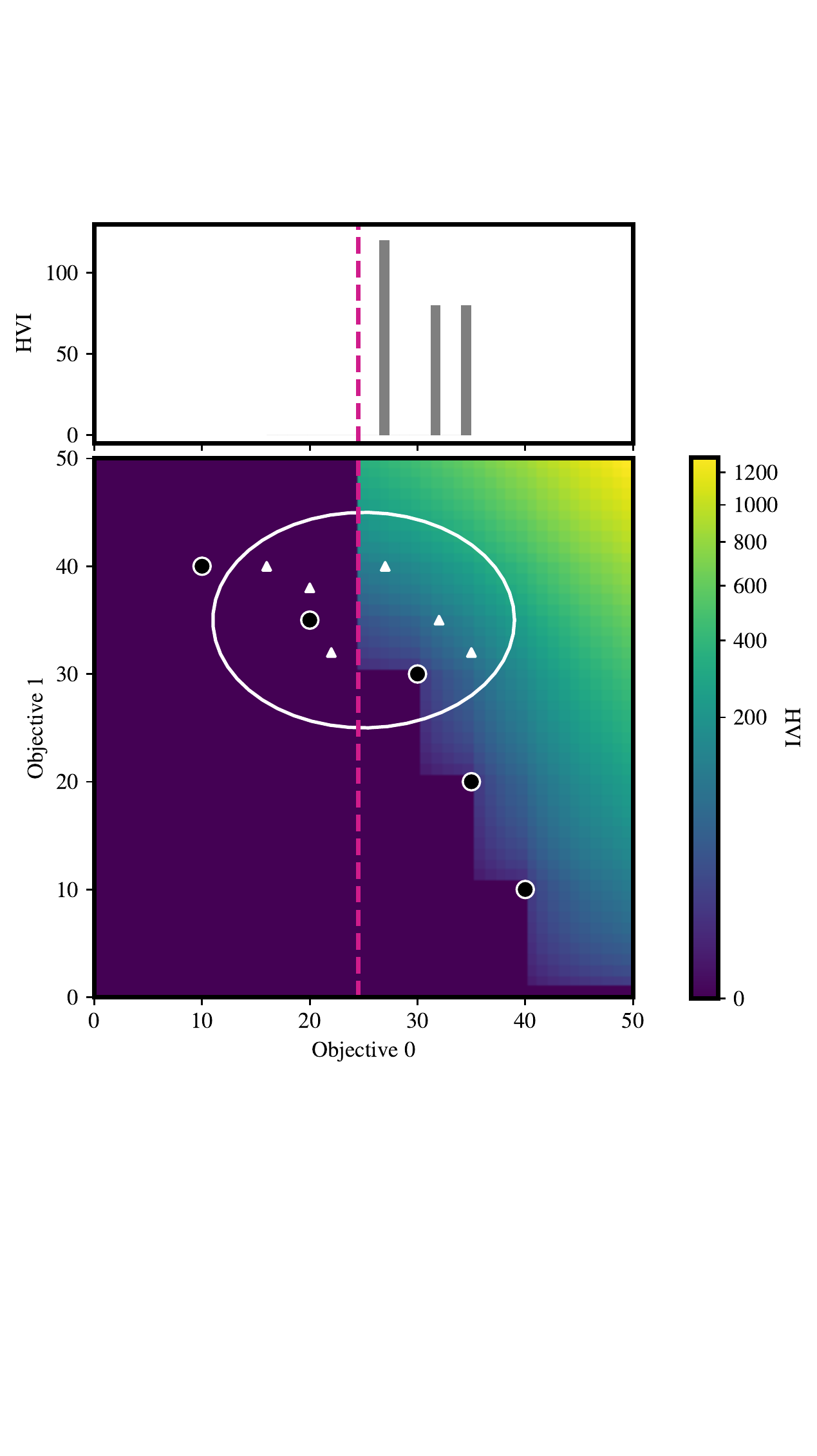} 
        \caption{PropertyDAG resampling} \label{fig:pareto_schematic_dag}
    \end{subfigure}
    \caption{Effect of resampling the surrogate posteriors on the acquisition function. Suppose the magenta dashed lines represent our threshold for Objective 0 such that we want to acquire candidates that maximize Objective 1 \textit{and} exceed this threshold in Objective 0. \textit{Bottom:} The black dots constitute our baseline Pareto front. Colors of the grid indicate HVI (Equation~\ref{eq:ehvi}) computed from each posterior sample at the given location in the objective space, in the (a) default and (b) PropertyDAG settings. Consider six samples (white triangles) from the posterior (white contour). PropertyDAG transforms the posterior samples below the threshold in Objective 0 such that their HVI contribution is zero. \textit{Top:} The HVI corresponding to each sample before (a) and after (b) the resampling.}
\end{figure}

\subsection{Zero-inflated modeling} 
\label{sec:zero_inflation} 

Biological properties tend to carry excess zeros, or null values. Their zero-inflated nature motivates us to employ statistical models that account for large incidences of zeros \cite{eggers2015statistical}. The zero-inflated negative binomial (ZINB) distribution has been applied to model discrete counts in single-cell RNA-seq data \citep{grun2014validation}. For each objective $y_k$, our zero-inflated model assigns a binary random variable $b_k \in \{0, 1\}$ to generate the zeros and $r_k \in \mathbb{R}$ to generate the remaining wide dispersion of continuous non-zero values.

Assume $f$ is non-negative (equivalently, that it is bounded from below).
Given a training dataset $\mathcal{D}_t$ available at time $t$, we decompose $p(\hat f_k \mid \mathcal{D}_t)$ as follows:
\begin{align} \label{eq:combined}
   p(\hat f_k(\bm x) = c \mid \mathcal{D}_t) = \begin{cases}
   p(b_k = 0 \mid \bm x, \mathcal{D}_t) & \text{if } c = 0, \\
   p(b_k = 1 \mid \bm x, \mathcal{D}_t) \ p(r_k = c \mid \bm x, \mathcal{D}_t, \theta_r) & \text{else},
   \end{cases}
\end{align}
where $p(r_k \mid \bm x, \mathcal{D}_t, \theta_r)$ is a regressor parameterized by $\theta_r$ trained to predict $f_k$ given $b_k = 1$.
For simplicity we have assumed $p(r_k=0|\bm{x}, \mathcal{D}_t, \theta_r)=0$.
Since Gaussian processes (GPs) are often used as surrogates for BO \citep[see, e.g.][for a review on GPs]{williams2006gaussian} and common GP assumptions fail for sparse, multi-modal data, separating out the non-zero mode of the data can improve posterior inference.

To accommodate objective hierarchy, each $p(b_k \mid \bm x, \mathcal{D}_t)$ decomposes further as
\begin{align} \label{eq:binary_posterior}
    p(b_k \mid \bm x, \mathcal{D}_t) = \begin{cases}
    0 & \text{if } \exists j \in {\rm pred}(k) \text{ s.t. } b_j = 0, \\
    p(b_k \mid \bm x, \mathcal{D}_t, \theta_b) & \text{else},
    \end{cases}
\end{align}
where $p(b_k \mid \bm x, \mathcal{D}_t, \theta_b)$ is a classifier parameterized by $\theta_b$ trained to predict $\mathds{1}\{f_k(\bm x) > 0\}$ and ${\rm pred}(k)$ are the predecessors, or ancestral nodes, in the PropertyDAG corresponding to property $k$.

This general framework, presented in terms of a zero-inflated, continuous-valued objective (a mixture of a delta function at zero and a continuous distribution), applies to binary-valued objectives and continuous-valued objectives without zero inflation, which can be viewed as specific cases taking $p(r_k|\bm{x}, \mathcal{D}_t, \theta_r) = p(r_k) = \mathcal{N}(0, \sigma^2)$ with very small $\sigma$ and $p(b_k=1 \mid \bm x, \mathcal{D}_t, \theta_b)=1$, respectively.

\subsection{Resampling} 
\label{sec:resampling}

Using a simple resampling trick, we modify the posterior samples from the surrogate models to enforce the parent-child relationships specified in PropertyDAG. As before, consider a property $y_k$ and refer to its predecessors, or ancestral nodes, as ${\rm pred}(k)$. Suppose we have drawn single a posterior sample and obtained $\beta_{k'} \sim  p(b_{k'}|\bm{x}) \in \{0, 1\}$ and $\rho_{k'} \sim  p(r_{k'}|\bm{x}) \in \mathbb{R}$ for each ${k'} \in \{1, \dotsc, K\}$. Without any modification, Equation~\ref{eq:combined} would yield the following sample $\gamma_k$ of $y_k$:
\begin{align} \label{eq:no_resampling}
{\gamma}_k = \begin{dcases*}
0 & if $b_k = 0$ \\
\rho_{k} & if $b_k = 1$.
\end{dcases*}
\end{align}

Instead, we begin at the top level and proceed down the levels of PropertyDAG to impose dependencies between $b_k$ and its predecessor properties $\{b_{k'}\}_{k' \in {\rm pred}(k)}$. If $y_k$ is a top-level property, then ${\rm pred}(k) = \emptyset$ and $\hat{\beta}_k = \beta_k$. Otherwise, $y_k$ has parent properties and we have
\begin{align} 
\label{eq:binary_resampling}
\hat{\beta}_k = \begin{dcases*}
0 & if $\exists j \in {\rm pred}(k) \text{ s.t. } b_j = 0$ \\
\beta_k  & else.
\end{dcases*}
\end{align}

We can then use the modified binary samples $\{\hat{\beta}_k\}_{k=1}^K$ to obtain our effective sample $\hat{\gamma}_k$ of $y_k$:
\begin{align} \label{eq:regress_resampling}
\hat{\gamma}_k = \begin{dcases*}
0 & if $\beta_k=0$ or $\exists j \in {\rm pred}(k) \text{ s.t. } b_j = 0$ \\
\rho_k & else.
\end{dcases*}
\end{align}

Let $\bm{\gamma} \coloneqq [ {\gamma}_1, \dotsc, {\gamma}_K ] \in \mathbb{R}^K$ and $\bm{\hat{\gamma}} \coloneqq [ \hat{\gamma}_1, \dotsc, \hat{\gamma}_K ] \in \mathbb{R}^K$ and denote the transformation at the sample level described in Equations~\ref{eq:binary_resampling} and \ref{eq:regress_resampling} as $h: \mathbb{R}^K \rightarrow \mathbb{R}^K$ such that $h(\bm{\gamma}) = \bm{\hat{\gamma}}$. 

We repeat this resampling procedure to other posterior samples. The transformed samples are then used to evaluate NEHVI (Equation~\ref{eq:nehvi}) via MC integration. More precisely, suppose we draw $S$ posterior samples in parallel for a design candidate $\bm{x}^*$ (reflecting both aleatoric and epistemic uncertainties) and the previously observed designs $\mathcal{X}_t = \{\bm{x}^{(n)}\}_{n=1}^{N_t}$ (reflecting the aleatoric uncertainty) and denote each draw as $\bm{{\gamma}}^*_s$ and $G_s \coloneqq \{\bm{{\gamma}}_s^{(n)}\}_{n=1}^{N_t}$, respectively, for $s=1, \dotsc, S$. Then the MC approximation of NEHVI can be efficiently evaluated as:
\begin{align} \label{eq:nehvi_mc}
    & \bm{\hat{\gamma}}_{s}^* = h(\bm{\gamma}_{s}^*), 
    \: \bm{\hat{\gamma}}_{s}^{(n)} = h(\bm{\gamma}_{s}^{(n)}) 
    \quad \forall s=1, \dotsc, S,
    \: \forall n=1, \dotsc, N_t \quad \nonumber \\
    & \alpha_{\rm NEHVI} \left( \bm{x}^* \right) \approx \frac{1}{S} \sum_{s=1}^{S} {\rm HVI} \left(\mathcal{P}^{* [s]}_t, \mathcal{P}^{[s]}_t |\bm{r}_{\rm ref} \right),
\end{align}
where $\mathcal{P}_t^{[s]} = \{\bm{{\gamma}}_s: \bm{{\gamma}}_s \in G_s, \: \nexists \: \bm{{\gamma}}_s' \in G_s \textit{ s.t. } \bm{{\gamma}}_s' \succ \bm{{\gamma}}_s \}$ and $\mathcal{P}^{* [s]}_t = \mathcal{P}^{[s]}_t \cup \{\bm{\hat{\gamma}}_{s}^*\}$.

Figures~\ref{fig:pareto_schematic_default} and \ref{fig:pareto_schematic_dag} illustrate the effect of the PropertyDAG resampling on each posterior sample's HVI contribution to EHVI. 

\section{Experiments} \label{sec:results}

We perform simulated active learning experiments on two synthetic tasks and a real-world antibody design task. We use NEHVI (Equation~\ref{eq:nehvi}) as our acquisition function and evaluate it via MC integration (Equation~\ref{eq:nehvi_mc}). In each experiment, we test three types of acquisitions: (1) batched, multi-objective BO with PropertyDAG (``qNEHVI-DAG''), (2) without PropertyDAG (``qNEHVI''), and (3) random. Our main metric is the number of acquired ``joint positive'' designs, designs that exceed the chosen thresholds in all objectives according to the specified PropertyDAG. We refer to the batch size as $q$. 

\subsection{Penicillin production dataset}
\label{sec:penicillin_experiment}

This task is based on the penicillin production simulator introduced in \cite{liang2021scalable}. We defined the goal as minimizing the CO$_2$ byproduct emission while ensuring that the fermentation time is below a set threshold and the yield exceeds a set threshold ($K=3, \mathcal{X}=\mathbb{R}^7$). We negated the latter two objectives to define a maximization problem and assume the PropertyDAG: $\{y_{0, 0}\} \rightarrow \{y_{1, 0}\} \rightarrow \{y_{2, 0}\}$, where $y_{0, 0}=$ Yield (``Objective 0''), $y_{1, 0}=$ Negative fermentation time (``Objective 1''), and $y_{2, 0}=$ Negative CO$_2$ byproduct (``Objective 2''). Zero-mean Gaussian noise was added to the input, following \cite{daulton2020differentiable}.

We fit an exact GP to model $r_k$ and an approximate GP with the variational evidence lower bound (ELBO) to model $b_k$, separately for each Objective $k$. We drew 512 posterior samples to evaluate qNEHVI. 

\begin{figure}[ht!] 
    \centering
    \begin{subfigure}[t]{0.45\textwidth}
        \centering
        \includegraphics[scale=0.35]{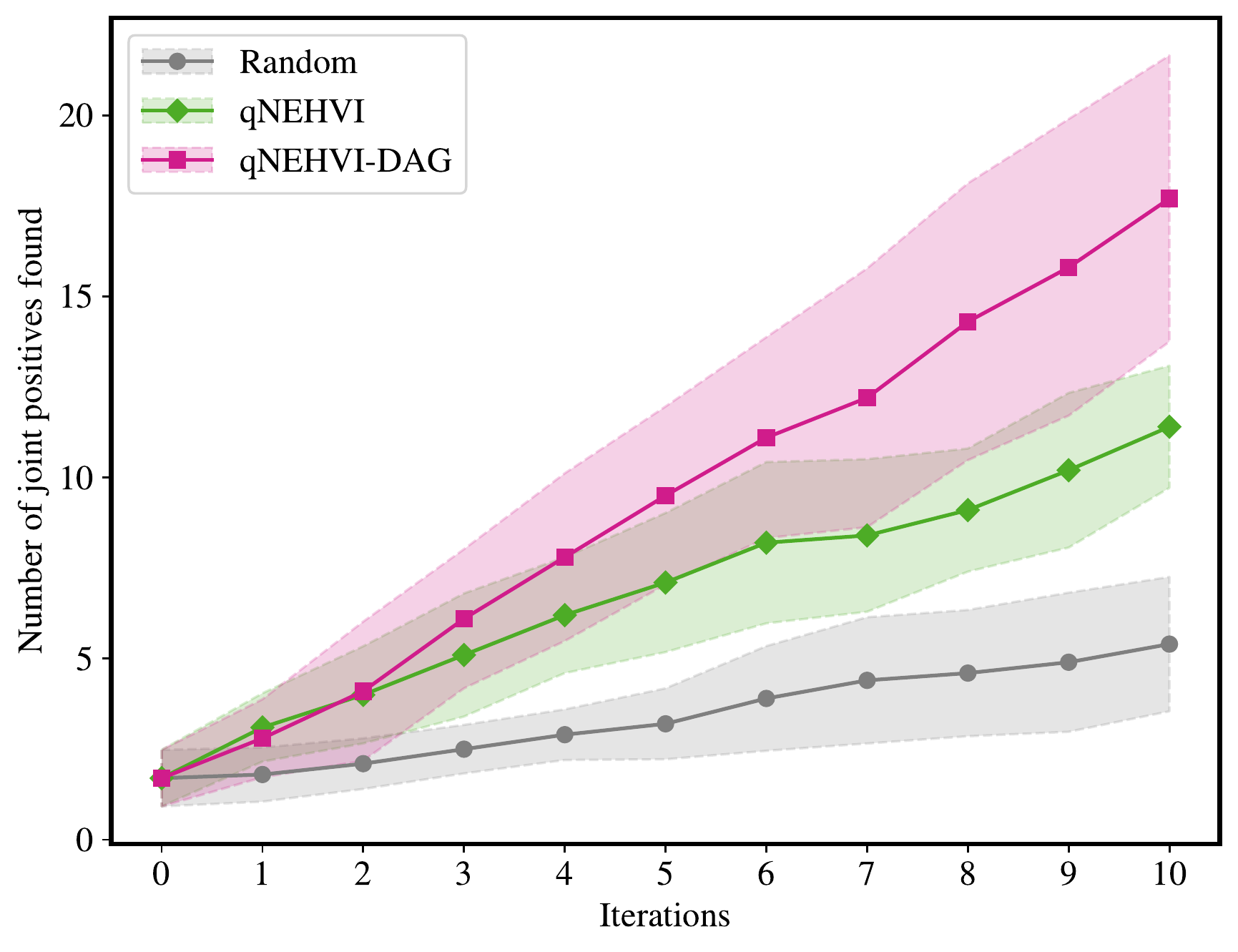}
        \caption{Penicillin production \label{fig:ab_log_p}}
        \label{fig:penicillin_joint_pos}
    \end{subfigure}
    \hfill
    \begin{subfigure}[t]{0.45\textwidth}
        \centering
        \includegraphics[width=\linewidth]{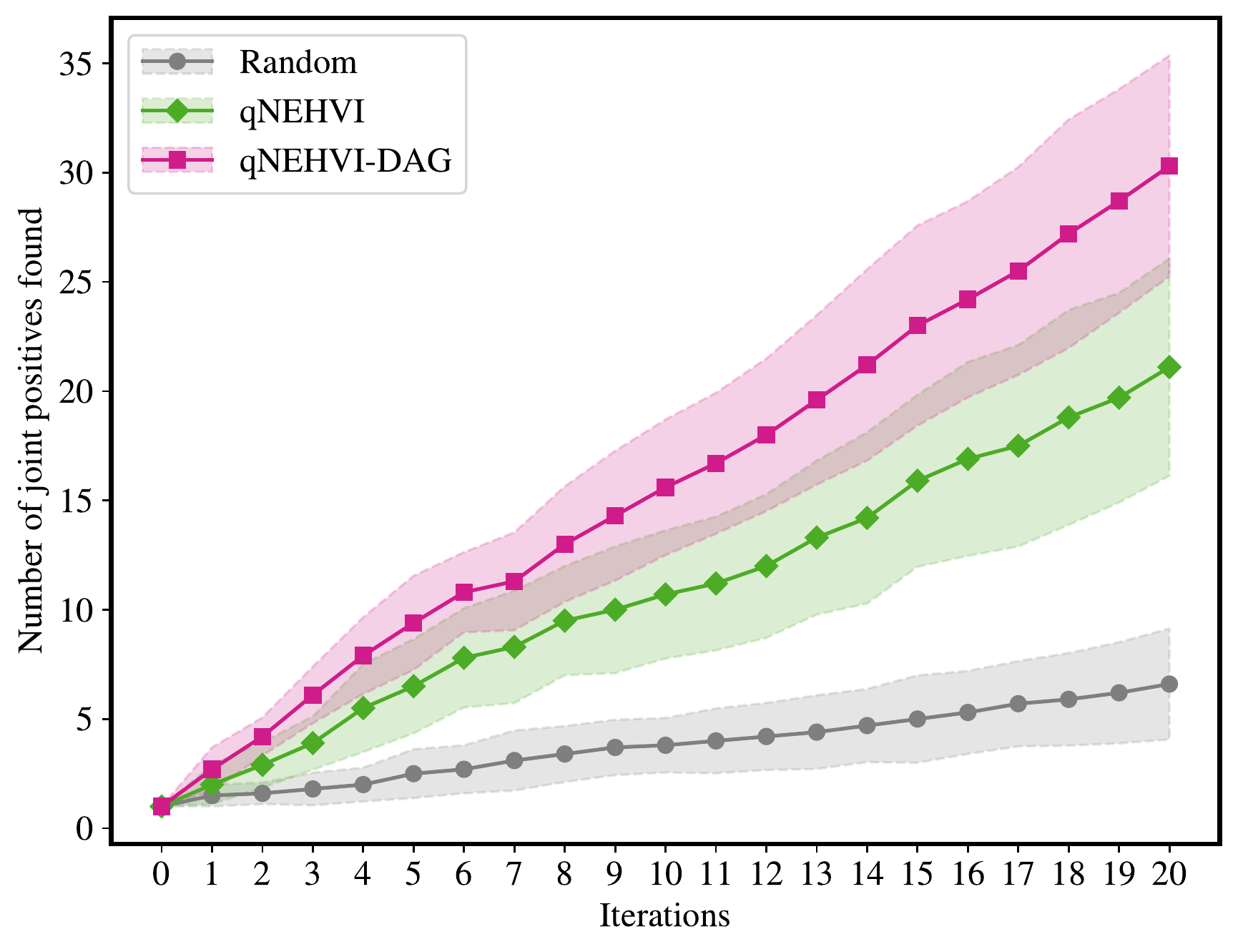}
        \caption{Branin-Currin toy problem} \label{fig:bc_joint_pos}
    \end{subfigure}
    \caption{Number of joint positives over simulated active learning iterations. Error bands are standard deviations over (a) five and (b) ten trials.}
    \vspace{-0.3cm}
\end{figure}

\begin{figure}[ht!]
  \centering
  \includegraphics[width=0.7\textwidth, trim=0.cm 1cm 0.cm 4cm]{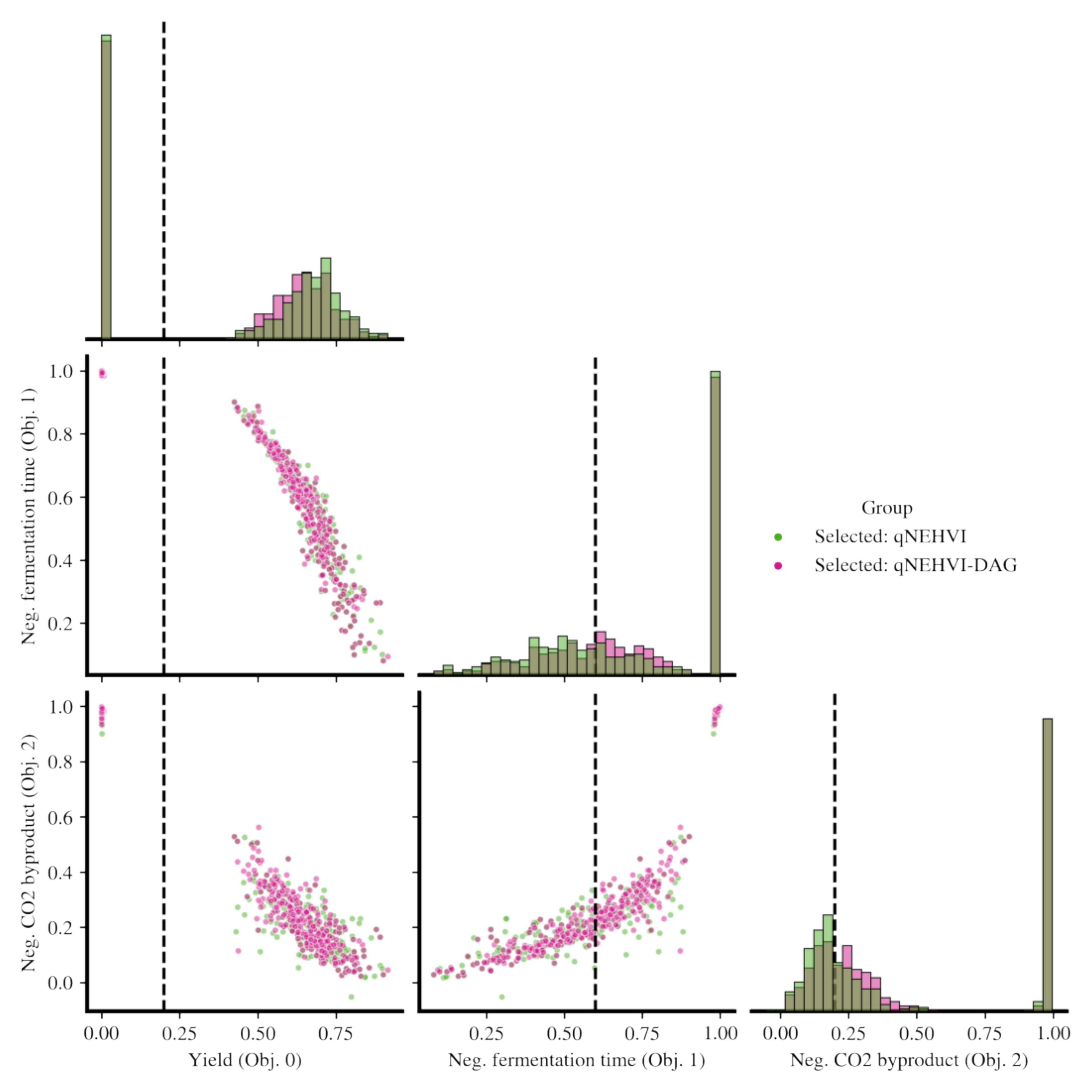}
  \caption{Pairwise Pareto front visualization for the penicillin production task showing the ground-truth objectives of final selections. We pool the selections over five trials. Black dashed lines are the thresholds we impose. \label{fig:penicillin_pareto}}
\end{figure}

We executed 10 rounds of simulated active learning by initializing the surrogates with 8 training points and selecting $q=4$ out of 80 randomly-sampled pool of candidate points in iteration. The three acquisition modes (qNEHVI-DAG, qNEHVI, and Random) were subject to the same initial training points and candidate pool each round. The entire experiment was repeated 5 times. Figure~\ref{fig:penicillin_joint_pos} shows that qNEHVI-DAG identifies significantly more joint positives than do qNEHVI and Random over active learning iterations. Figure~\ref{fig:penicillin_pareto} compares the qNEHVI and qNEHVI-DAG selections for every pair of objectives, for the final (after Iteration 10) selections stacked across the 5 repeated trials. For every objective, qNEHVI-DAG identifies more examples to the right of the threshold (black dashed lines) than do qNEHVI and Random. 


\subsection{Branin-Currin toy problem}
\label{sec:bc_experiment}

This task is based on an analytic Branin-Currin test function from \cite{daulton2020differentiable} with $\mathcal{X}=\mathbb{R}^2$ and $K=2$. We reformulated this task to simulate the antibody design task (Section \ref{sec:ab_design_experiment}) in a controlled environment. We defined the PropertyDAG, $\{y_{0, 0}\} \rightarrow \{y_{1, 0}\}$, where $y_{0, 0}=$ Dimension 0 (``Objective 0'') and $y_{1, 0}=$ Dimension 1 (``Objective 1''). Objective 0 was transformed into binary values using a set threshold. Objective 1 was zero-inflated and real-valued. Posterior inference was performed following the same procedure described in Section \ref{sec:penicillin_experiment}.

\begin{figure}[ht]
  \centering
  \includegraphics[width=\textwidth]{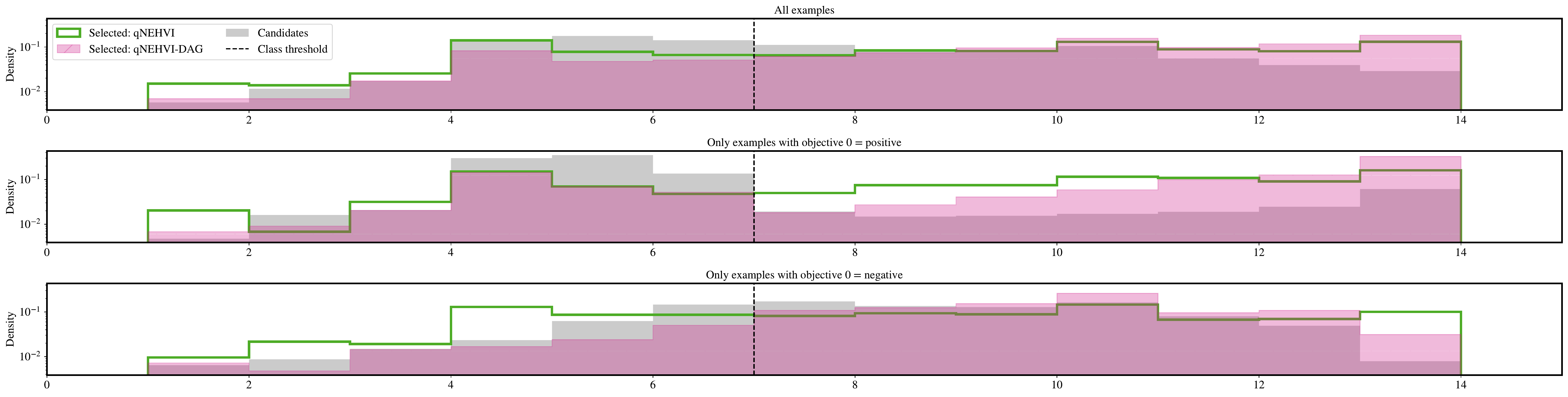} 
  \caption{Distribution of final selected candidates in the Branin-Currin problem. We stack the selections over 10 repeated trials. \textit{Top}: all selections. \textit{Middle}: selections for which Objective 0 = 1.  \textit{Bottom}: selections for which Objective 0 = 0. \label{fig:bc_obj}}
\end{figure}

We executed 20 rounds of simulated active learning by initializing the surrogates with 6 training points and selecting $q=4$ out of 40 randomly-sampled pool of candidate points in iteration. The entire experiment was repeated 10 times. Figure~\ref{fig:bc_joint_pos} shows that qNEHVI-DAG identifies significantly more joint positives than do qNEHVI and Random over active learning iterations. Figure~\ref{fig:bc_obj} compares the qNEHVI-DAG, qNEHVI, and Random selections for Objective 1, for the final selections stacked across the 10 repeated trials. Overall, qNEHVI-DAG identifies more examples to the right of the threshold (black dashed lines) than do qNEHVI and Random, and the improvement is more pronounced for the identification of joint positives (middle panel). 

\subsection{Antibody design}
\label{sec:ab_design_experiment}

The antibody design task is derived from real-world dataset of antibody sequences and their measured \textit{in vitro} properties, affinity and expression. As in the toy problem (Section~\ref{sec:bc_experiment}), we defined the PropertyDAG, $\{y_{0, 0}\} \rightarrow \{y_{1, 0}\}$, where $y_{0, 0}=$ Expression (``Objective 0'') and $y_{1, 0}=$ Affinity (``Objective 1''). Objective 0 was binary-valued, i.e. expressing or not. Objective 1 was zero-inflated and real-valued.

\begin{figure}[ht] 
    \centering
    \begin{subfigure}[t]{0.45\textwidth}
        \centering
        \includegraphics[scale=0.35]{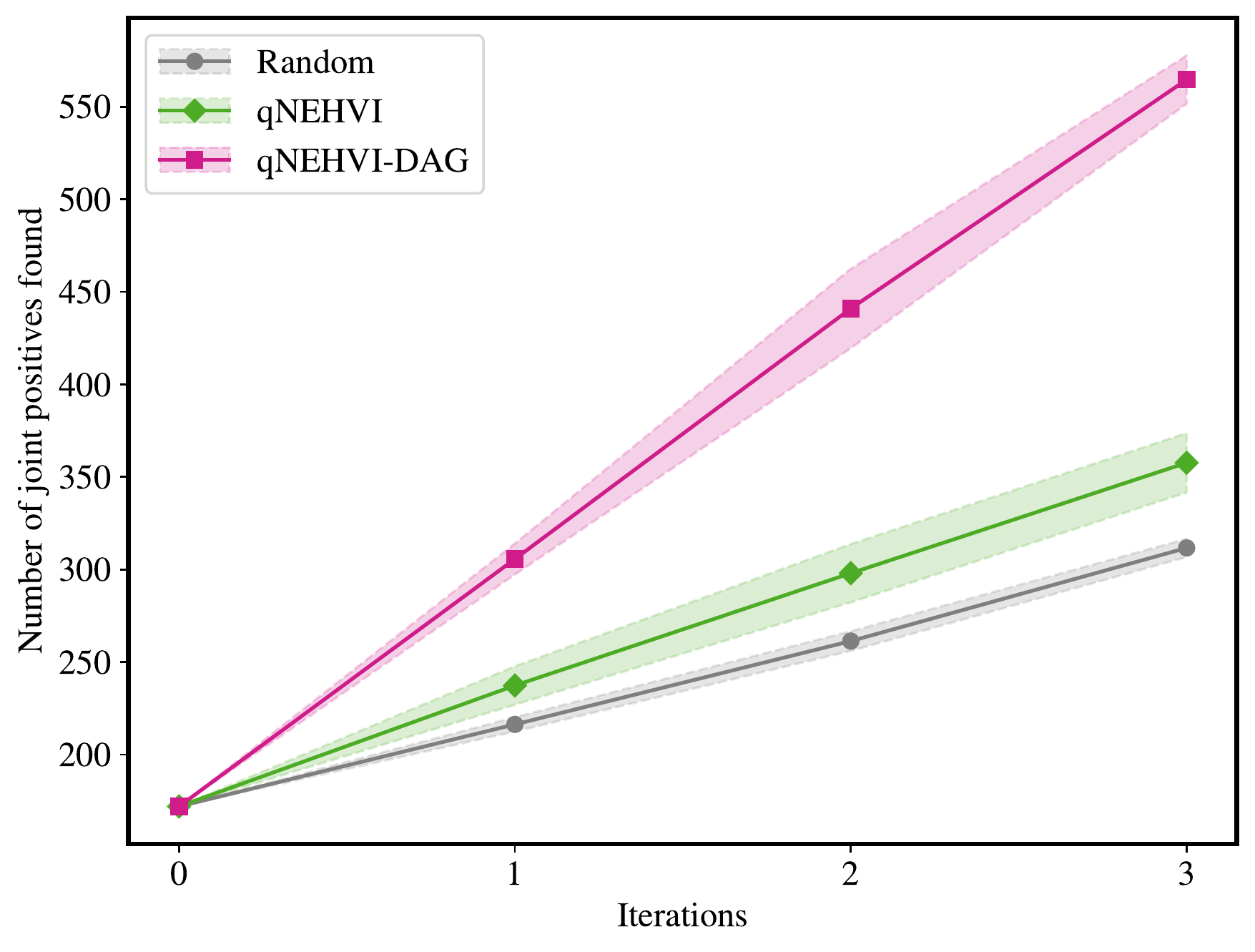}
        \caption{Joint positives identified}
        \label{fig:ab_joint_pos_over_iters}
    \end{subfigure}
    \hfill
    \begin{subfigure}[t]{0.45\textwidth}
        \centering
        \includegraphics[width=\linewidth]{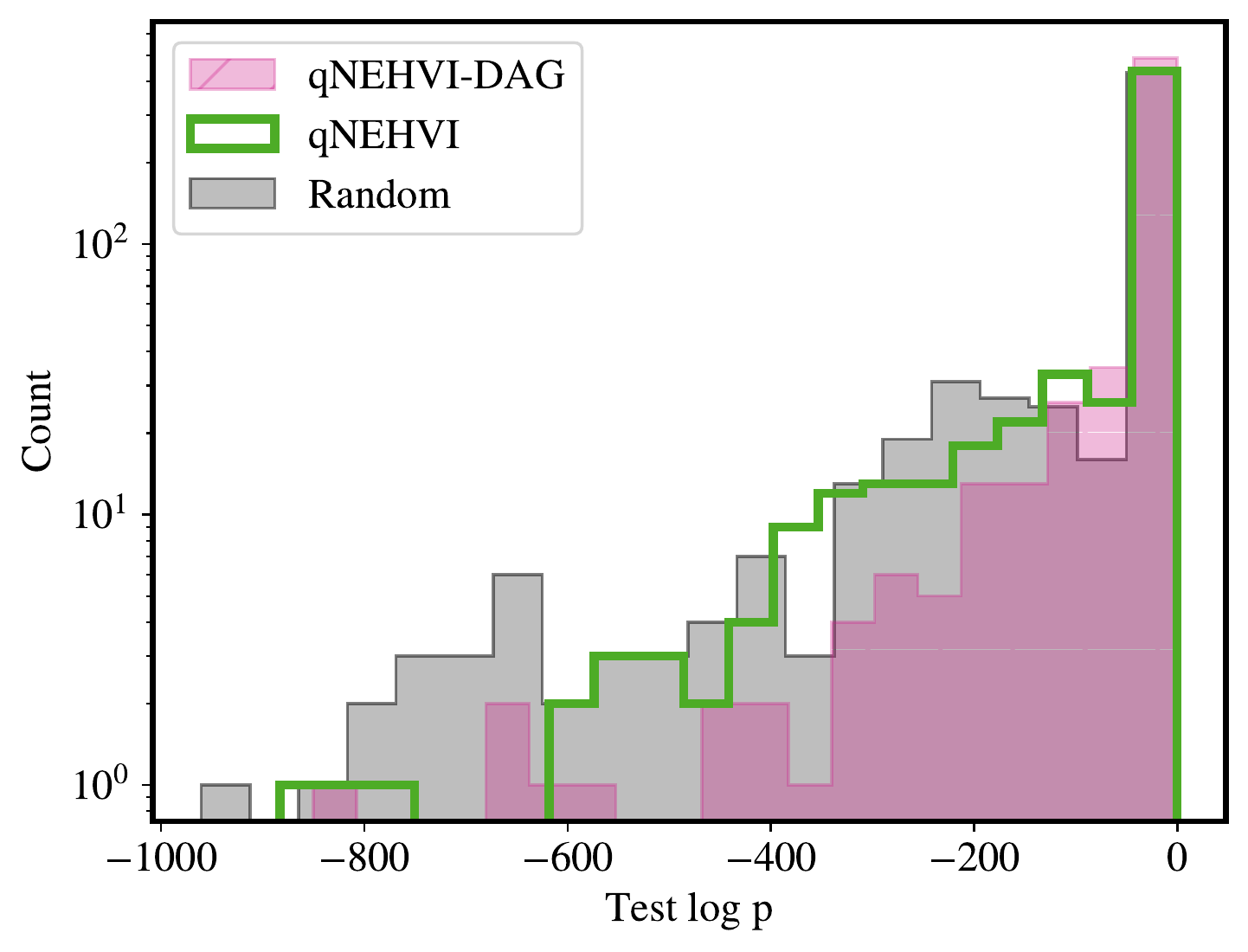}
        \caption{Log posterior density for joint positives} \label{fig:ab_log_p}

    \end{subfigure}
    \caption{(a) Number of joint positives (expressing binders) for the antibody design problem. Error bands are standard deviations over three data splits for qNEHVI-DAG and qNEHVI and additionally over five repeated trials for each data split for Random. (b) Log posterior density on binding affinity evaluated at the lab affinity measurements ($\log p$) for the test set consisting of 600 datapoints. Each $\log p$ value is averaged over five data splits. \label{fig:ab_log_p}}
    \vspace{-0.3cm}
\end{figure}

We executed 3 iterations of simulated active learning and repeated the entire procedure 5 times. To simulate active learning, we split the entire dataset of 4,022 variable-length protein sequences, designed as antibodies for an anonymized target antigen A, into 5 groups of sizes 1230, 736, 746, and 600. The first group served as the initial training set for the surrogates, the following three groups as the ``candidate pools'' from which we selected 200 candidates in each iteration, and the last served as a held-out test set. As shown in Figure~\ref{fig:ab_joint_pos_over_iters}, qNEHVI-DAG once again outperforms qNEHVI and Random in the number of joint positives. The log posterior density evaluated at the affinity measurements for the joint positives (expressing binders) in the test set is also highest for qNEHVI-DAG, which indicates that the surrogate models from qNEHVI-DAG had the most accurate beliefs about the joint positives after the final iteration.

\section{Conclusion}

Our proposed method, PropertyDAG, sits on top of the existing multi-objective BO framework to make it amenable to a common scenario in drug design, where a hierarchical structure, or partial ordering, exists among the objectives. It modifies the surrogate posteriors so that each objective is modeled as zero-inflated (a mixture of excess zeros and a continuous distribution) and parent properties in PropertyDAG are prioritized before the children. Empirical evaluations shows that PropertyDAG-BO can identify significantly more designs that are jointly positive (i.e. exceeding a chosen threshold in all properties) than does standard BO. By encapsulating our experimental and biological priors on the relationship between molecular properties, our method promises to accelerate computational drug discovery.

\medskip



\clearpage

{
\small
\bibliographystyle{unsrtnat}
\bibliography{main}
}

\appendix




\end{document}